\documentclass[sigconf]{acmart}

\usepackage{booktabs} % For formal tables
\usepackage{multirow}
\usepackage{lipsum}
\usepackage{subcaption}

\setcopyright{rightsretained}
\acmDOI{10.475/123_4}

\acmISBN{123-4567-24-567/08/06}

\acmConference[SIGKDD'17]{Workshop on Machine Learning for Creativity}{Aug 2017}{Halifax - Canada} 
\acmYear{2017}
\copyrightyear{2017}

\begin{document}
\title{Story Generation from Sequence of Independent Short Descriptions}
%\titlenote{Produces the permission block, and copyright information}
%\subtitle{Extended Abstract}
%\subtitlenote{The full version of the author's guide is available as \texttt{acmart.pdf} document}

\author{Parag Jain$^\dagger$ \hspace{0.1cm} Priyanka Agrawal$^\dagger$ \hspace{0.1cm} Abhijit Mishra$^\dagger$ \hspace{0.1cm} Mohak Sukhwani$^\dagger$\\ Anirban Laha$^\dagger$ \hspace{0.1cm} Karthik Sankaranarayanan$^\dagger$\\
$^\dagger$IBM Research India,\\
\{pajain06,priyanka.agrawal,abhijimi,mosukhwa,anirlaha,kartsank\}\\@in.ibm.com}
% \affiliation{%
%   \institution{IBM Research}
%   \city{Bangalore} 
%   \country{India}
% }
% \email{pajain06@in.ibm.com}

% \author{Priyanka Agrawal}
% \affiliation{%
%   \institution{IBM Research}
%   \city{Bangalore} 
%   \country{India}
% }
% \email{priyanka.agrawal@in.ibm.com }

% %\author{Lars Th{\o}rv{\"a}ld}
% %\authornote{This author is the
% %  one who did all the really hard work.}
% \author{Abhijit Mishra}
% \affiliation{%
%   \institution{IBM Research}
%   \city{Bangalore} 
%   \country{India}
% }
% \email{abhijimi@in.ibm.com}

% \author{Mohak Sukhwani}
% \affiliation{%
%   \institution{IBM Research}
%   \city{Bangalore} 
%   \country{India}
% }
% \email{mosukhwa@in.ibm.com}

% \author{Anirban Laha}
% \affiliation{%
%   \institution{IBM Research}
%   \city{Bangalore} 
%   \country{India}
% }
% \email{anirlaha@in.ibm.com}

\renewcommand{\shortauthors}{P. Jain et al.}

\begin{abstract}
Existing Natural Language Generation (\textsc{nlg}) systems are weak AI systems and exhibit limited capabilities when language generation tasks demand higher levels of creativity, originality and brevity. Effective solutions or, at least evaluations of modern \textsc{nlg} paradigms for such creative tasks have been elusive, unfortunately.
This paper introduces and addresses the \textbf{task of coherent story generation from independent descriptions}, describing a scene or an event. Towards this, we explore along two popular text-generation paradigms -- (1) Statistical Machine Translation (\textsc{smt}), posing story generation as a translation problem and (2) Deep Learning, posing story generation as a sequence to sequence learning problem. In SMT, we chose two popular methods such as phrase based SMT (\textsc{pb-SMT}) and syntax based SMT (\textsc{syntax-SMT}) to `translate' the incoherent input text into stories. We then implement a deep recurrent neural network (\textsc{rnn}) architecture that encodes sequence of variable length input descriptions to corresponding latent representations and decodes them to produce well formed comprehensive story like summaries. The efficacy of the suggested approaches is demonstrated on a publicly available dataset with the help of popular machine translation and summarization evaluation metrics. 
%Our general observation is that while SMT based approaches are good at finding appropriate lexical replacements for the input text (thereby improving evaluation scores), deep learning framework does a better job in generating coherent and fluent stories. 
We believe, a system like ours has different interesting applications- for example, creating news articles from phrases of event information.

\end{abstract}

\maketitle

\section{Introduction}
Recent advances in machine learning based approaches for natural language generation have led to exploration of many diverse but related text generation tasks. However, the existing systems/ approaches can be classified as weak~AI systems. According to the classical definition \cite{kurzweil2005singularity}, a strong AI based \textsc{nlg} system should perform language generation in the same manner, expressing similar levels of creativity, originality and brevity as humans. We proceed towards building such systems for the difficult task of automatic story generation that demands the above mentioned human qualities. 

\begin{figure}[t]
\begin{center}
   \includegraphics[width=\linewidth]{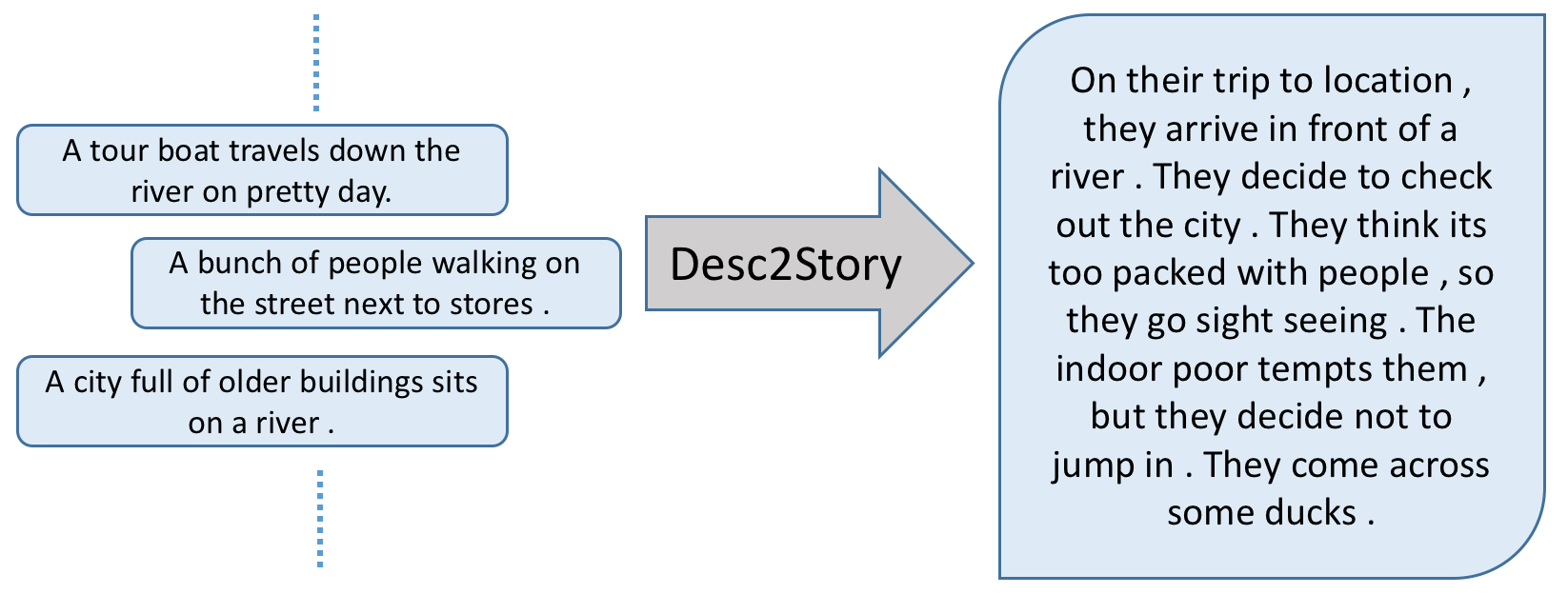}
\end{center}
\caption{Our system generates variable length stories for variable length input descriptions. The standalone textual descriptions describing a scene or event are converted to human like coherent summaries.}
\label{fig:abstract}
\end{figure}

\begin{figure*}[t]
\begin{center}
\includegraphics[width=0.6\linewidth]{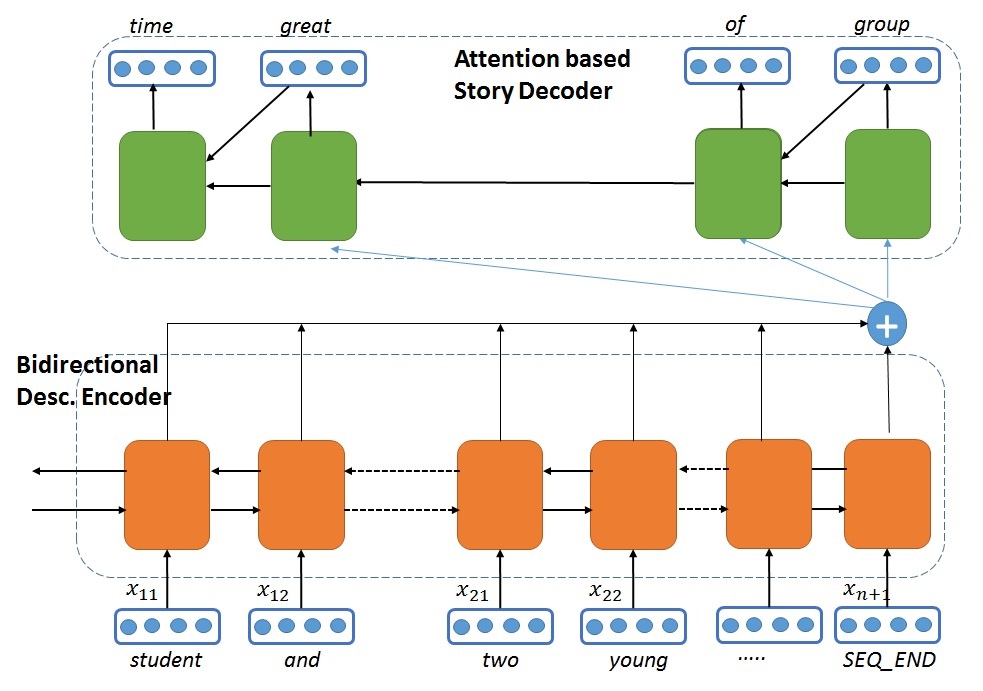}
\end{center}
\caption{Sequence to Sequence Recurrent Neural Network Architecture for Desc2Story. Orange and green colored blocks represent GRU cells in encoder and decoder respectively.}
\label{fig:propArch}
\end{figure*}

We introduce the task of generating coherent narratives form a sequence of independent short descriptions, as shown in Figure~\ref{fig:abstract}. The standalone descriptions (in general), although are sufficiently informative enough to describe a scene or an event, they lack the flavor of temporal context and human touch, viz. describing a Christmas celebration scene as \textit{`The church was all ready for the big day. The candles were lit and people were arriving. In came the alter boys. They began to read from the book. They put the lights out and the candle were light'} is far more apt and intriguing than describing it by independent sentences - `\textit{people are in a large candlelit room in which there is a Christmas tree.}', `\textit{ritual candles are lit on stands in a darkened room.}', `\textit{a priest is holding a special ceremony.}' \textit{etc.} 

The proposed system of story generation is framed as a sequence-to-sequence neural machine learning network that learns latent representations of the input descriptions to generate output summaries. The success of ~\textsc{rnn}s to model sequential language tasks have made it a de-facto choice of ~\textsc{nlp} community. From character level language models~\cite{seqOCR} to task of machine translations~\cite{nmt},~\textsc{rnn}s have outperformed all of its present day contemporaries. We envision an end-to-end ~\textsc{rnn} system that understands the context holistically and generates cohesive textual stories. The input to our system is a set of short textual descriptions, each describing a scene or an event and the system produces semantically rich human like stories. Figure ~\ref{fig:abstract} demonstrates a working example.

Our work is motivated by two independent considerations -- (1) Introduce the concept of `creativity' for the machine generated text, i.e. a set of same input descriptions should have ability to generate varied story-line when parameterized over user choice of `theme'. (2) Leverage underlying ephemeral cues in input text to generate contextually relevant and coherent output stories. We take a step towards this by proposing a new computational regime of creative text content generation. We draw in our parallels from neural language generation modeling and more specifically sequence-to-sequence recurrent neural network, ~\textsc{seq2seq}. Unlike phrase Based~\cite{pbt} and syntax based~\cite{smt} ~\textsc{smt} which translate words and phrases independently within a sentence, ~\textsc{seq2seq} considers entire sentence as a unit to be translated. We leverage ~\textsc{seq2seq} architecture \textit{i.e.}, deep recurrent neural network (\textsc{rnn}) with attention mechanism for the problem of coherent story generation. Additionally, we also pose story generation as a translation problem and report results for both phrase based and syntax based SMT towards this. 

We provide an outline of all explored approaches and our network design choices in Section~\ref{Section:Approach}. This is followed by detailed Sections ~\ref{Section:Experiment}-\ref{Section:Discussion} describing our experimental setup, results and analysis respectively on publicly available dataset. Lastly we describe avenues where the proposed schema can be utilized along with conclusions and future directions in Section~\ref{Section:Conclusion}.

\begin{table*}
  \centering
    \begin{tabular}{l ccc cc c }
    \toprule
     & \textbf{\# Docs} & \multicolumn{2}{c}{\textbf{Avg. \# Sents}} & \multicolumn{2}{c}{\textbf{Avg. \# Words}} & \textbf{Avg. \# non-overlapping Words}\\ 
     & & Caption & Story & Caption & Story & (excluding stop words)\\
    \midrule
    Train & 32987 & 5 & 6 & 52 & 56 & 23\\
    Val & 4168 & 5 & 6  & 51 & 57 & 22\\
    Test & 4145 & 5 & 6 & 51 & 56 & 23\\
    \bottomrule
    \end{tabular}
    \caption{Statistics of the dataset used for experimentation}
    \label{tab:datastats}
\end{table*}

\section{Related Work}
\label{Section:Review}
Text based story generation as a cognitive task has seen considerable traction over last many years. The onslaught of deep learning architectures and advancements in hardware infrastructure has made this task much more relevant in today's time. Image captioning~\cite{pan2004automatic, you2016image, ordonez2011im2text}, video understanding~\cite{Yao_2015_ICCV, guadarrama2013youtube2text, BMVC2015_117}, narratives in virtual environments~\cite{aylett1999narrative} to novellas by computers~\cite{meehan1976metanovel,barrie2014computers} are all manifestations of present day interest ~\textsc{ai} community has over text based story generating creative systems.

Computational models to mimic human creativity (here: ~\emph{Story Generation}) are not new. In past storytelling has been approached as an analytical activity of discovering best fact to present to the listener~\cite{pemberton1989modular}. In an attempt to add engagement, reflections and make stories user centric, more sensitive models modifying stories with human feedback were proposed~\cite{bailey1999searching, perez2001mexica}. The traditional approaches were succeeded by much more sophisticated virtual environments with multi agents framework keeping tracks of plots, narratives, user interest etc. ~\cite{aylett1999narrative, theune2003virtual, gervas2005story}. Text based interactive narratives are the other kind of storytelling variations that have been tried in past. Multitude of approaches ranging from evolutionary (genetic) algorithms~\cite{mcintyre2009learning} to mining of crowd sourced information~\cite{swanson2008say, li2013story} from web have also been explored.

Attempts to generate cohesive comprehensible stories have not gone beyond short text snippets. Recent works on Visual storytelling ~\cite{huang2016visual, LiuFMC16} aim at generating story by harnessing rich cognitive information from a sequence of images. In this paper, we introduce and address a much tougher setting  -- starting from incoherent description, this is the first known work that discuss the prospects of converting it to human like stories. Our setting is deprived of the rich contextual information that is available in the form of images in traditional  Visual storytelling ~\cite{huang2016visual}. 
%The high amount of information that is latent instead of directly observable (complementary image information) is far more in present setting which makes the present problem much more difficult to solve.
Such a system, however, has many interesting applications- for example, creating news articles from phrases of event information. 

This paper is a preliminary study on story generation from textual input and explores the effectiveness of two traditional language generation paradigms. Our experiment section presents both qualitative and quantitative analyses of the different approaches in the respective paradigms. 
%approaches is also presented in the experiment section.

%We present a preliminary study and propose a system to achieve such a task. A comparative analysis with respect to the traditional state of art approaches is also presented in the experiment section. 

% We take a first step towards this direction -- starting from incoherent descriptions, this is the first known work that discuss the prospects of converting it to human like stories. 

\section{Explored Approaches for Story Generation}~\label{Section:Approach}
\begin{table}[b]
  \centering
    \begin{tabular}{l l l l l}
    \toprule
    \textbf{Method} & \textbf{BLEU-4} & \textbf{METEOR} & \textbf{TER} & \textbf{ROUGE-L} \\
    \midrule
    \textsc{pb-SMT} & 3.50 & 10.30 & 102.95 & 0.179\\
    \textsc{syntax-SMT} & 3.40 & 10.06 & 102.03 & 0.180\\
    \textsc{Seq2Seq (50)} & 1.63 & 0.07 &  89.38 & 0.160\\
    \textsc{Seq2Seq (128)} & 1.84 & 0.07 & 89.35 &  0.163\\
    \textsc{Seq2Seq (256)} & 1.98 & 0.07 & 89.23 & 0.166\\
    \bottomrule
    \end{tabular}
    \caption{Evaluation results for \textsc{smt} and \textsc{seq2seq} methods. For each \textsc{seq2seq} method, embedding dimensionality is mentioned in brackets.}
    \label{tab:results}
\end{table}
Story generation from independent textual-descriptions is the process of transforming one form of text into another. Towards this, we explore along two popular text-generation paradigms -(1) Statistical Machine Translation (\textsc{SMT}), posing story generation as a translation problem and (2) Deep Learning, posing story generation as a 
sequence to sequence learning problem. In \textsc{SMT}, we chose two popular methods 
such as phrase based SMT and syntax based \textsc{SMT} where as in sequence to sequence learning, we implement different variants of recurrent neural networks (\textsc{RNN}), 
empowered with attention mechanism.

\subsection{Phrase based Statistical Machine Translation \textsc{pb-SMT}}
Phrase based Statistical Machine Translation (\textsc{pb-SMT}) deals with (a) finding best possible target language phrase-map for phrases in the given source text (referred to as prediction of alignment) and (b) combining the phrase-maps together to synthesize the target language text with the objective to maximize fluency (referred to as decoding). 
In our setting incoherent captions are treated as \emph{source} and the corresponding generated stories in the used data-sets are used as \emph{target}. 
In \textsc{pb-SMT}, for learning phrase alignment, we use multi-threaded GIZA++ \cite{gao2008parallel} and decoding is done using the MOSES decoder \cite{koehn2007moses}. The development-set is used to tune the \textsc{pb-SMT} system using MERT mechanism \cite{och2003minimum}.

\subsection{Syntax based Statistical Machine Translation \textsc{syntax-SMT}}
Syntax-based translation (\textsc{syntax-SMT}) is based on the idea of translating syntactic units, rather than single words or strings of words (as in phrase-based MT), 
i.e. (partial) parse trees of sentences/utterances \cite{yamada2001syntax}. For the task of story-generation, we use similar training configuration as \textsc{pb-SMT} to learn 
syntactic unit correspondences; for decoding the \emph{chart-decoder} of moses \cite{koehn2007moses} is used for syntax-based (tree-to-tree based) target-sentence generation.
\subsection{Sequence-to-sequence Recurrent Neural Network}
As an initial approach for story generation from independent descriptions, we use sequence-to-sequence recurrent neural network (\textsc{seq2seq}) architecture \cite{seq2seq} popularly used for machine translation \cite{nmt}, question answering, summarization and other text generation tasks. 
We use a bidirectional encoder that processes the set of independent descriptions separated by a delimiter and \textsc{seq\_end} marking the end of set. This encoding of descriptions is given to an attention enabled decoder that generates the story sequence. Refer to Figure \ref{fig:propArch} for an overview.

\begin{figure*}[t]
\begin{subfigure}{0.49\textwidth}
\includegraphics[width=\linewidth]{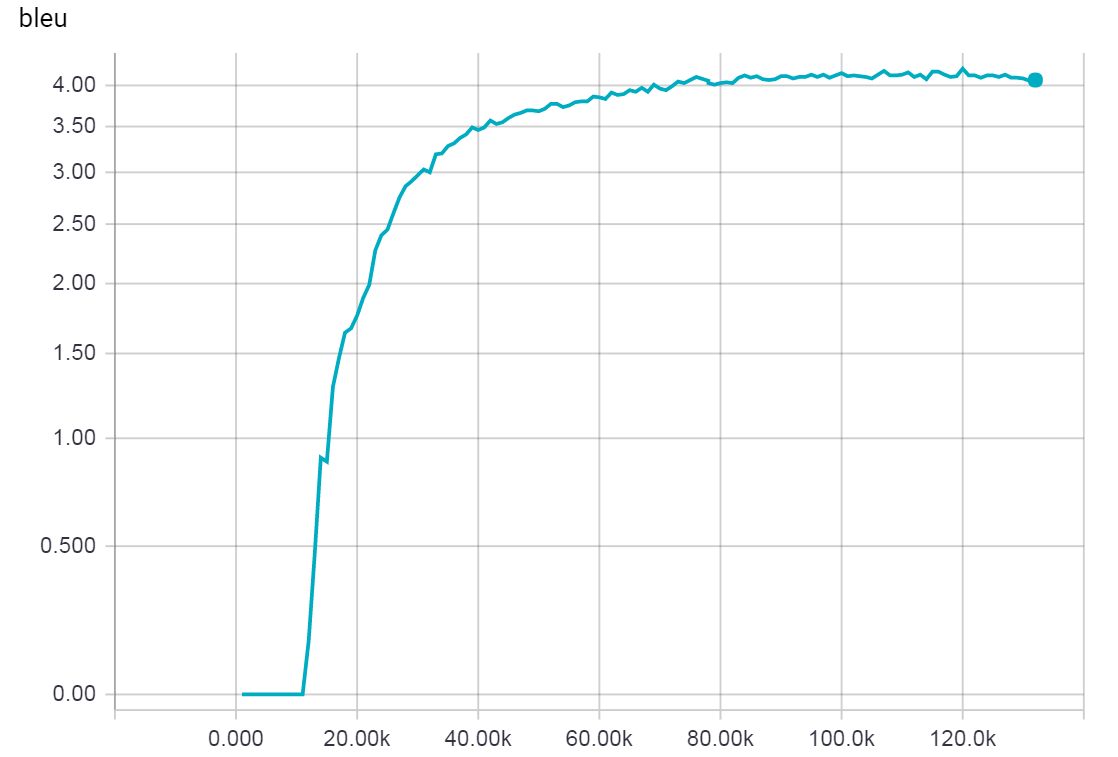}
\caption{BLEU score} \label{fig:Results_A}
\end{subfigure}
\hspace*{\fill} % separation between the subfigures
\begin{subfigure}{0.49\textwidth}
\includegraphics[width=\linewidth]{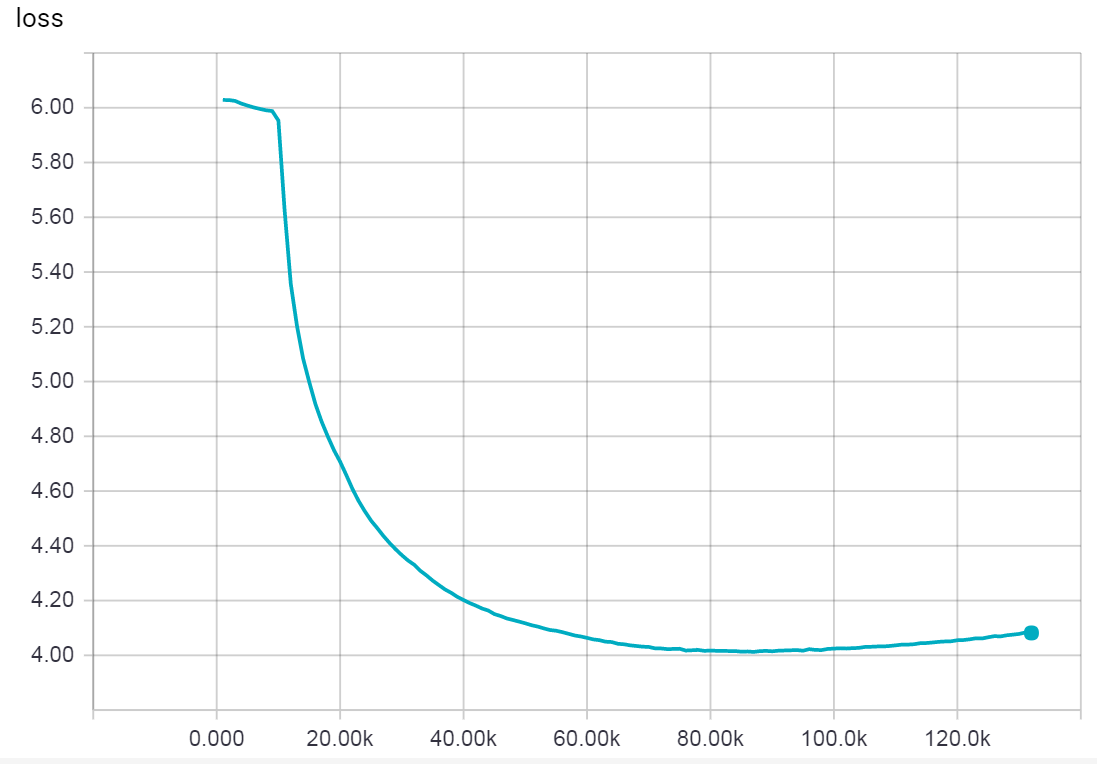}
\caption{Training loss} \label{fig:Results_B}
\end{subfigure}
\caption{Graphs depicting changes in BLEU score and Loss over the iterations during training.} \label{fig:Results}
\end{figure*}

\section{Experiment Setup}
\label{Section:Experiment}
We now describe the experimental setup below, sharing details on the dataset and configuration details of the story generation models. 
%Experiments for SMT based system were implemented using GIZA++ and MOSES toolkit. For seq2seq experiments we use Tensorflow \textsc{seq2seq} \citet{Britz:2017} implementation and trained it on a single Nvidia K80 GPU. 

\subsection{Dataset}

In the absence of availability of the any dataset for specific task, we use `Text Annotations' from recently introduced Visual Storytelling Dataset~\cite{huang2016visual} (VIST) for our experiments. Barring the images, we use both `Descriptions of Images-in-Isolation (DII)' and `Stories of Images-in-Sequence (SIS)' for our experiments. In all, there are $41300$ image sequences aligned with caption and story pairs. The text data is crowd sourced using Amazon Mechanical turk -- participant\'s were shown image sequences from the dataset to generate text descriptions for captions and stories. We believe these pairs act as a good (input-output) proxy for our setup as both of them describe a `scenes' represented by image sequences. Independent captions describe each image as a separate event and stories are a coherent version of text describing the image sequence as a whole. Few sample input sets of descriptions and output stories are listed as a part of Table \ref{tab:examples}. 

VIST dataset is split into training, validation and test sets and Table \ref{tab:datastats} summarizes various statistics of these sets like average number of words and sentences per caption and story. 
Usually the stories are longer and on an average have one sentence more compared to set of captions. More that 50\% of words in a story are different from that in the captions. In fact, on an average, ~$41$\% of the story are non-stop words unseen in the input, thus making the task of generating the precise story tougher given limited set of training inputs. 

\begin{table*}[t]
\begin{center}
\begin{tabular}{|p{3cm}|p{4.25cm}|p{4.25cm}|p{4.25cm}|}
\hline
{\bf Input Descriptions} & a student and an older man sit down to talk about technology. & group of older people gathered for bingo night in hall. & the leaf has begun to change it `s color . \\
  \multirow{4}{0em}{} &  two young women speaking with an older man wearing a suit.  &
 a girl sitting at a table eating pancakes with fruit. 
&  close up of a bush with small pink flower clusters. \\
& man with gray hair and eyeglasses talking to woman in cap sitting at a computer. & a group of historical reenactors torment tourists who had hoped to have a fun vacation . & purple and white flowers growing outdoors together .\\
&  a young man and an older man are sitting in front of a monitor while referencing a book. & a line of people walking through the woods.& \\
& group of students standing for picture in classroom. &  & \\
\hline
{ \bf \textsc{pb-SMT} } & the family is so proud of him . the man is talking to her friends . this one has improved flash . i gave the speech . he is welcome to the graduation . we had a great time and we love him for a book . at the end of the day , and we had a great time . one of the class . & everyone arrived for the bingo hall . at the end of the night . we had a great dinner . more fruits . all of the history here . our reenactors torment was lovely . we took a walk through the forest . & in the woods . this begun to it . it is a clusters . the flowers were beautiful flowers and plants .\\
\hline
{ \bf \textsc{syntax-SMT} } & the family is so proud of him . he has improved flash . to get to the fun . he and his . to my eyes . i turned on and my dad . we played in the world . the students love to read a book . and we had a great time . & the family gathered at the end of the night . there was a bingo hall . at t     he end of the night . we ate breakfast at the end of the ruins . the fruits      of our torment reenactors our lovely . we took a walk through the woods .
& this begun in the woods . it was to support . there was a nice . the flowers and plants . the flowers were in clusters photos .\\
\hline
{ \bf \textsc{SEQ2SEQ (256)} } & a group of friends met for a meeting . they talked about the plans . they talked about the company 's company . they had a great time at the meeting . everyone had a great time .
&
a group of friends went on a trip to location . they went to a historical site . they saw many interesting buildings . they also had a map to eat . they had a great time .
& we decided to take a trip to the botanical gardens . there was a beautiful view of the water . we also saw some lily pads . there was a lot of flowers on the ground . we also saw some lily pads .\\
\hline
\textbf{Ground Truth} & the gentleman sat with the boy to discuss the diagrams . he then asked the young ladies if they needed help with anything . he spoke to the man about his interest in technology . he then met with this fellow , to discuss his future plans in engineering . the students took a picture to remember the special day . &
we went to the syrup festival. the kids got to eat pancakes. there was also a demonstration on how syrup is made. we got a tour of the woods. and it ended in the production lab.
&   i have been outside doing micro photography for a class project . love how it is possible to get a good blur in pictures like this . these flowers wer e so beautiful . one of my class subjects is nature . nothing beats getting out and taking pictures of sites like this . most people never experience this . last but not least a single red flower . this day of shooting turned out very good .\\
\hline
\end{tabular}
\end{center}
\caption{Example Case studies on VIST dataset}
\label{tab:examples}
\end{table*}
%\subsection{Baseline: Statistical Machine Translation (SMT)}
%~\cite{smt}

\subsection{Model Details}
For SMT related experiments, we use the MOSES toolkit \cite{koehn2007moses}. For phrase based translation model learning,\emph{grow-diag-final-and} heuristic is opted and to tackle lexicalized reordering, we use the \emph{msd-bidirectional-fe} model. GIZA++ is configured to apply the principles of IBM model 4 and 5 \cite{brown1993mathematics} and the HMM alignment model \cite{och2000improved} for alignment learning. The assumption behind selecting the higher order IBM models is that we expect the possibility of addition of foreign words as well as dropping of words when a given text is transformed into a story, and higher order IBM models are good at tackling such nuances. 

We tuned the trained SMT models using Minimum Error Rate Training (MERT) with default parameters (100 best list, max 25 iterations). We trained a 5-gram language model using the target side (stories) of the training data. For this, we use use Kneser-Ney smoothing algorithm implemented in KenLM toolkit \cite{heafield2011kenlm}. Batch training of multiple SMT systems was done using the Moses Job Scripts \footnote{\url{https://bitbucket.org/anoopk/moses_job_scripts}} experiment management system. We use similar training and tuning configuration for setting up the syntax based SMT system. The default rule-learning mechanism of MOSES is opted for training by setting the \texttt{hierarchical} flag in the moses training script. Decoding during tuning and testing is carried out using the moses chart decoder. 

For sequence to sequence learning, we implement Desc2Story framework used in our experiments. Fig 2 provides an overview of the model, our framework is based on sequence to sequence encoder decoder architecture including attention mechanism at the decoder \cite{bahdanau2014neural}. Encoder is a single layer bidirectional Gated Recurrent Unit(GRU) \cite{cho2014learning} including dropout. Decoder is a 2 layer GRU using attention mechanism at each time step. We have used Adam optimizer to optimize the cross entropy loss function. Both encoder and decoder has a dropout \cite{srivastava2014dropout} with probability $p = 0.8$. We tried 50, 128 and 256 dimensional RNN at encoder and decoder. During testing for SEQ2SEQ model, we use beam search with beam-width set to $5$. Batch size of 32 was used while training.

\subsection{Evaluation Metrics}
For evaluation, we rely on popular Machine Translation evaluation metrics such as BLEU-4 \cite{blue}, METEOR \cite{meteor}, and TER \cite{ter} and ROUGE-L \cite{rouge}, frequently used for evaluating summarization output. While these evaluation metrics may not adequately capture all necessary aspects of story generations, in the absence of alternative evaluation methods specific to story generation, these metrics could provide first a level insight into the applicability of popular \textsc{nlg} paradigms towards a difficult task like story generation.

\section{Results and Discussion}
\label{Section:Discussion}
In this work, we attempted to try a few off-the-shelf techniques to solve a complex task as generating a story from simple textual descriptions in a sequence, without the taking another input or modality. The evaluation results are reported in Table \ref{tab:results}, clearly suggesting that the 
overall scores are not very high. This indicates the limitations of basic off-the shelf methods for a highly creative task such as story generation even though they work well for simpler machine translation, paraphrase generation or document summarization tasks. This is a very complex task as stories can be generated with a greater diversity in content compared to the above tasks. While the narrative generation is itself a grand challenge, metrics to evaluate such a task is also inadequate.  It is well-known that BLEU and other metrics have their shortcomings even in simpler natural language generation settings as they are mostly based on n-gram match. This is another reason why scores of our baseline models are low even though we found coherent enough outputs as shown in Table \ref{tab:examples}. 

From Table \ref{tab:results}, it is clear that SMT systems produce higher BLEU/ METEOR/ROUGE values with \textsc{pb-SMT} being the best. \textsc{Seq2Seq} scores, on the other hand, suffer as \textsc{seq2seq} models are embedding-based approaches which can lead to less scores on $n$-gram based match evaluation even though they are able to produce better readable and coherent stories. This is evident from the anecdotal examples in Table \ref{tab:examples}. This is supported also by the fact that TER, which is not based on  n-gram match, is lower in case of \textsc{seq2seq} indicating better translation quality by lowering the translation edit rate. Figures \ref{fig:Results_A} and \ref{fig:Results_B} show how the training of \textsc{seq2seq} model saturates with the iterations (each iteration is a new batch),  suggesting that the model has indeed converged.

Referring to the examples as in Table \ref{tab:examples}, we find that all methods are doing well to generate a story which is readable and coherent. This shows the models are capable enough to capture the grammatical aspects of a story based on the training data used. However, all models fall short of generating a story which is semantically related to the input descriptions, therefore, highlighting the need for more sophisticated models for creative story generation. We believe, our current models are too simplistic to realize beyond the co-occurrence statistics in the output, thus, unable to figure out the semantic relatedness between the  input and output. This could possibly be tackled to some extent by \emph{pretraining} the encoder and decoder in the \textsc{seq2seq} model using widely available data in the form of essays and novels. Another promising alternative is modeling the same problem with hierarchical recurrent neural networks.
%Another interpretation is that we have tried out techniques which rely on co-occurrence statistics in the output and this set of experiments show that we need to go beyond that. 
%One step in that direction can be using pretraining of encoder and decoder in the \textsc{seq2seq} model using widely available data in the form of essays and novels. 
%This may possibly alleviate the semantic relatedness problem between input and output to a good extent. 
However, in the long run, this kind of task also calls for novel generation of story at test time not just relying on mapping between input and output words/phrases learnt at training time. Thus, to achieve the distant goal of novel story generation which is somewhat semantically connected with the input, we have to look beyond just translation based techniques, which essentially maps input words to output words in sequence.

% {\color{green}  However, this is the first known attempt at this tough problem and we feel this is a step in the right direction.}

% \section{Discussion}
% \label{Section:Discussion}

% \subsection{Challenges in Story Generation}\label{Section:Challenges}

% Narrative generation is by far one of the Achilles heel of present day artificial intelligence.

% - SMT: Better lexical replacement but Seq2Seq, better readable / coherent stories. 

% - Need to go beyond basic models in SMT and Seq2Seq. May need to optimize eval metrics than simple losses (BLEU converges; explain figure 3 here). 

% - Metrics not adequate. 

% - Linguistic aspects of story generation - requirements of a good story. 

\section{Conclusion and Future Work}
\label{Section:Conclusion}
We introduced the task of automatically generating stories from independent one-liners. The scope of this task can eventually be expanded to generating stories from fewer input parameters like theme, actors, etc. This task is challenging not only because of insufficiency of information compared to the task attempted in VIST dataset but also because of the fact that the possible output space of stories is large. This is aggravated by the unavailability of good metrics to evaluate the approaches tried out.
Here we explored off-the-shelf sequence generation methods like  statistical machine translation and sequence to sequence neural networks as preliminary approaches towards solving this problem.

A future direction is to design trainable metrics for evaluating stories holistically to include aspects on creativity, coherency, novelty and other parameters compared to current score computation which is based on exact match. Creating an appropriate dataset for textual story generation is another important direction that we would like pursue.

%As the name suggests, Visual Story Telling (VIST) dataset was originally designed for images to story generation. Captions for an image capture the concept in the image partially which makes it challenging to stitch them into the story. Creating an appropriate dataset is another important direction that one can pursue. 

\bibliographystyle{ACM-Reference-Format}
\bibliography{storytelling-bib}

\end{document}